\documentclass[acmtog,nonacm]{acmart}

\usepackage{booktabs}
\usepackage{symbols}

\setcitestyle{nosort,square} %

\usepackage{hyperref}
\usepackage{amsmath}
\usepackage{graphicx}
\usepackage{comment}
\usepackage{kbordermatrix}
\usepackage{multirow,bigdelim}

\usepackage{array}
\usepackage{wrapfig}

\usepackage{csvsimple}
\usepackage{adjustbox}
\usepackage[percent]{overpic}

\definecolor{amber}{rgb}{1.0, 0.75, 0.0}
\definecolor{applegreen}{rgb}{0.55, 0.71, 0.0}
\definecolor{darkgoldenrod}{rgb}{0.72, 0.53, 0.04}
\definecolor{firebrick}{rgb}{0.7, 0.13, 0.13}

\newcommand{\ourmethod}{TextDeformer}

\usepackage{pgfplots}
\pgfplotsset{compat=newest}
\usepgfplotslibrary{groupplots}
\usepgfplotslibrary{dateplot}

\usepackage[bottom]{footmisc}
\usepackage[ruled,vlined]{algorithm2e}

\SetCommentSty{mycommfont}

\usepackage{bbold}
\usepackage{makecell}

\def\hlinewd#1{%
\noalign{\ifnum0=`}\fi\hrule \@height #1 %
\futurelet\reserved@a\@xhline} 
\renewcommand\footnotetextauthorsaddresses[1]{}

\begin{document}

\title{TextDeformer: Geometry Manipulation using Text Guidance}

\author{William Gao}
\affiliation{\institution{University of Chicago, USA}}
\author{Noam Aigerman}
\affiliation{\institution{Adobe Research, USA}}
\author{Thibault Groueix}
\affiliation{\institution{Adobe Research, USA}}
\author{Vladimir G. Kim}
\affiliation{\institution{Adobe Research, USA}}
\author{Rana Hanocka}
\affiliation{\institution{University of Chicago, USA}}

\begin{abstract}
 We present a technique for automatically producing a deformation of an input triangle mesh, guided solely by a text prompt. Our framework is capable of deformations that produce both large, low-frequency shape changes, and small high-frequency details. Our framework relies on differentiable rendering to connect geometry to powerful pre-trained image encoders, such as CLIP and DINO. Notably, updating mesh geometry by taking gradient steps through differentiable rendering is notoriously challenging, commonly resulting in deformed meshes with significant artifacts. These difficulties are amplified by noisy and inconsistent gradients from CLIP. To overcome this limitation, we opt to represent our mesh deformation through Jacobians, which updates deformations in a global, smooth manner (rather than locally-sub-optimal steps). Our key observation is that Jacobians are a representation that favors smoother, large deformations, leading to a global relation between vertices and pixels, and avoiding localized noisy gradients. Additionally, to ensure the resulting shape is coherent from all 3D viewpoints, we encourage the deep features computed on the 2D encoding of the rendering to be consistent for a given vertex from all viewpoints. We demonstrate that our method is capable of smoothly-deforming a wide variety of source mesh and target text prompts, achieving both large modifications to, e.g., body proportions of animals, as well as adding fine semantic details, such as shoe laces on an army boot and fine details of a face.

\end{abstract}

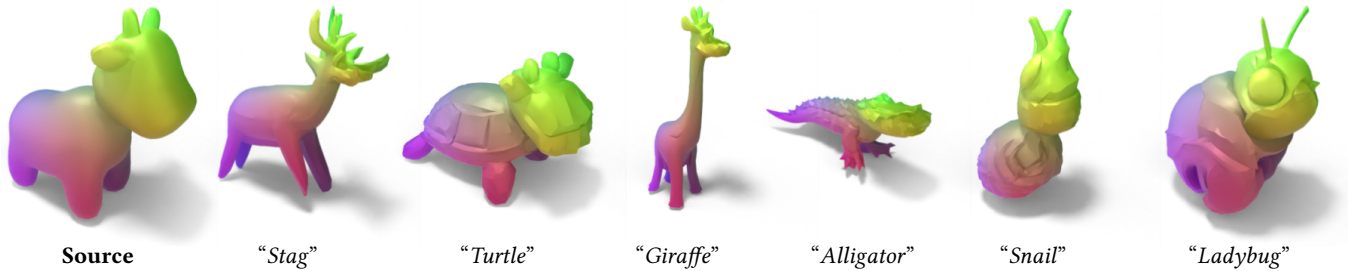
\begin{teaserfigure}
    \centering
    \newcommand{\pl}{-1}
    \begin{overpic}[width=\textwidth]{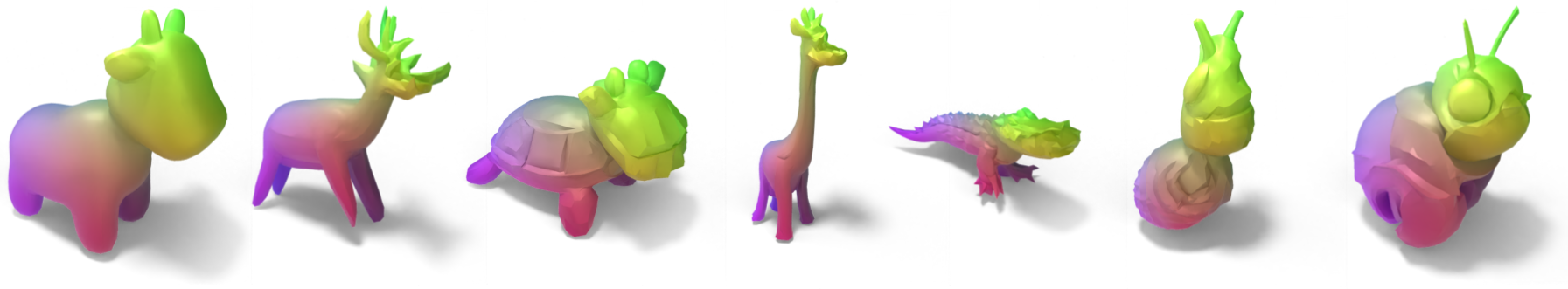} 
    \put(4.5,  \pl){\textcolor{black}{\textbf{Source}}}
    \put(19,  \pl){\textcolor{black}{``\textit{Stag}''}}
    \put(34,  \pl){\textcolor{black}{``\textit{Turtle}''}}
    \put(47,  \pl){\textcolor{black}{``\textit{Giraffe}''}}
    \put(60,  \pl){\textcolor{black}{``\textit{Alligator}''}}
    \put(74,  \pl){\textcolor{black}{``\textit{Snail}''}}
    \put(88,  \pl){\textcolor{black}{``\textit{Ladybug}''}}
    \end{overpic}
    \vspace{-1pt}
    \caption{\ourmethod{} deforms a source shape into various text-specified targets. The mesh colors visualize the smoothness of the mappings.}
    \label{fig:spot}
\end{teaserfigure}

\maketitle
\section{Introduction}
\vspace{-0.05cm}
This paper proposes a method to deform 3D meshes into other shapes through text guidance. Deforming meshes is a highly-researched problem in computer graphics and geometry processing, with applications in content creation~\cite{iwires}, character posing~\cite{jacobson_deformations}, and morphing~\cite{kraevoy_corrs_morphing}. Most existing techniques provide a user with the ability to control a deformation through control handles~\cite{jakab2021keypointdeformer, shechter2022neuralmls} which expose a space of coarse, low-frequency deformations. Such deformations are often referred to as detail-preserving. 
However, 3D modeling often also requires incorporating geometric details, where an artist needs to meticulously add details in a laborious process.

In this work, we aim to automate the entire deformation process, in order to automatically deform the mesh from its initial shape into the desired target shape, while preserving semantic correspondence between the source and the final shape.
To achieve this goal, we  follow the recent success of text-guided generative methods for images~\cite{dalle, styleclip}, meshes~\cite{text2mesh, clipmesh}, and NeRFs~\cite{dreamfusion}, by leveraging language as an intuitive tool for deforming shapes. Similarly to these previous works, our formulation does not rely on 3D training data, but instead leverages differentiable rendering to connect powerful pre-trained image encoders (such as CLIP~\cite{clip}) to provide a signal for modifying the geometry. After the deformation process, the resultant geometry respects the structure and characteristics of the source mesh, while visually adhering to the text specifications. 
In contrast to previous text-guided works which aim to either \emph{hallucinate} geometry from scratch~\cite{dreamfields, dreamfusion, sjc} or preserve the geometry of an input mesh~\cite{text2mesh} while adding detail, we instead focus on the \emph{shape deformation task}. 

Our framework manipulates an existing input shape, to enable producing high-quality geometry from the source mesh. Moreover,as can be seen in \figref{fig:spot}, our framework is capable of producing both low-frequency shape changes and high-frequency details (\eg the cow's neck is elongated when deforming to a giraffe) and incorporate details (scales are added when deforming to an alligator). The resulting correspondences from source shape to target are continuous and semantically meaningful (``leg deforms to leg''), which we visualize by coloring the source mesh (\eg in \figref{fig:spot} and throughout). This property is especially critical for shape-morphing applications.
\begin{figure}[t]
    \centering
    \begin{overpic}[width=\columnwidth]{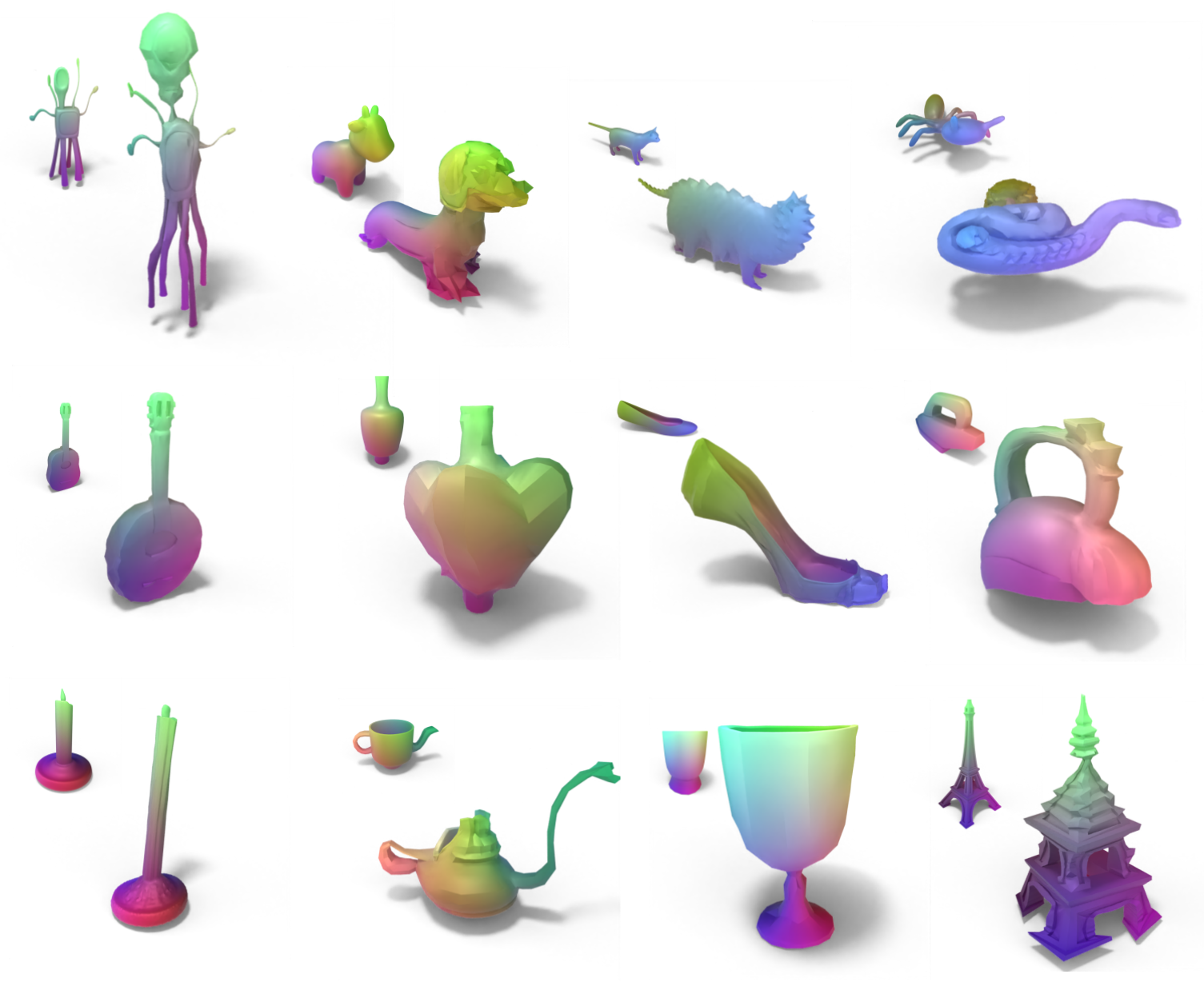}
    \put(5,  53){\textcolor{black}{``\textit{scary alien}''}}
    \put(27,  53){\textcolor{black}{``\textit{dachshund}''}}
    \put(53,  53){\textcolor{black}{``\textit{fluffy cat}''}}
    \put(79,  53){\textcolor{black}{``\textit{snake}''}}
    \put(7, 27){\textcolor{black}{``\textit{banjo}''}}
    \put(30, 27){\textcolor{black}{``\textit{heart vase}''}}
    \put(56, 27){\textcolor{black}{``\textit{high heel}''}}
    \put(83, 27){\textcolor{black}{``\textit{kettle}''}}
    \put(3, 0.5){\textcolor{black}{``\textit{incense stick}''}}
    \put(29, 0.5){\textcolor{black}{``\textit{genie lamp}''}}
    \put(56, 0.5){\textcolor{black}{``\textit{royal goblet}''}}
    \put(84, 0.5){\textcolor{black}{``\textit{pagoda}''}}
    \end{overpic}
    \caption{\textbf{Gallery of results.} Source meshes inset and target text below.}
    \label{fig:gallery}
\end{figure}
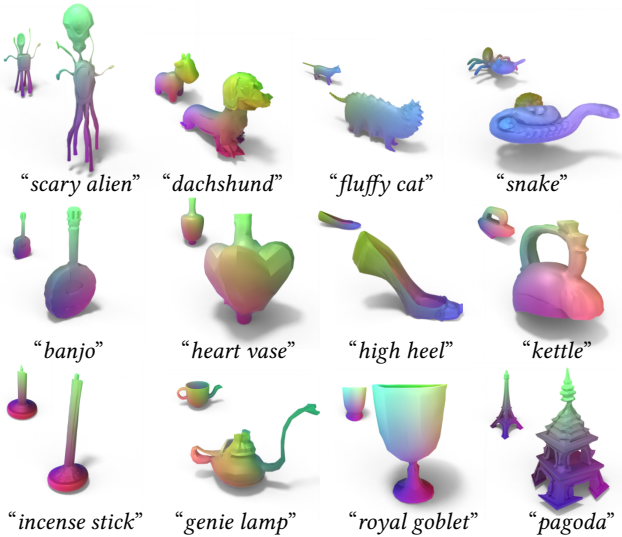

Thus, our framework is required to satisfy several properties: 1) produce high-quality surface geometry, with minimal self-intersections and noisy normals; 2) produce plausible results which match the text description; 3) adhere to the input geometry (e.g., deform the source's head into the target's head and not into body). 
\\This leads to several challenges which we solve through our technical contributions.

\begin{figure}[t]
    \centering
    \begin{overpic}[width=\columnwidth]{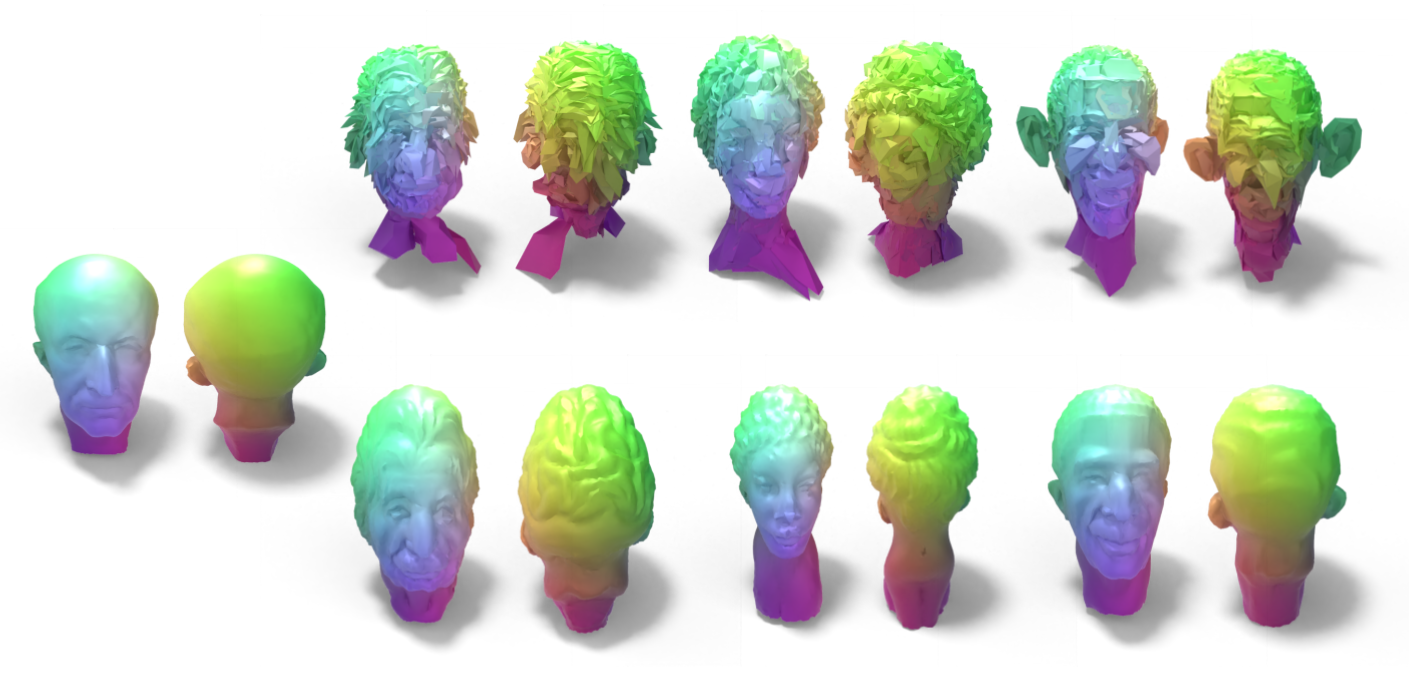}
    \put(7.5,  9){\textcolor{black}{\textbf{Source}}}
    \put(23.5,  -1.5){\textcolor{black}{``\textit{Albert Einstein}''}}
    \put(50,  -1.5){\textcolor{black}{``\textit{Bust of Venus}''}}
    \put(79.5,  -1.5){\textcolor{black}{``\textit{Obama}''}}
    \put(52,  22){\textcolor{black}{\textbf{Jacobians}}}
    \put(43,  46){\textcolor{black}{\textbf{Vertex Displacement}}}
    \end{overpic}
    \caption{\textbf{Globally-Coherent Deformations}. Max Planck deforms into different targets using our method (bottom row), where the front/back view is shown in pairs of left/right. Removing Jacobians and predicting displacements (top row) takes locally sub-optimal steps, which results in distorted shapes with significant artifacts. Jacobians produce global deformations, resulting in cleaner geometries and even prevents the spurious face mirroring (for example, in ``\textit{Obama}'').}
    \label{fig:global_jacobian}
\end{figure}
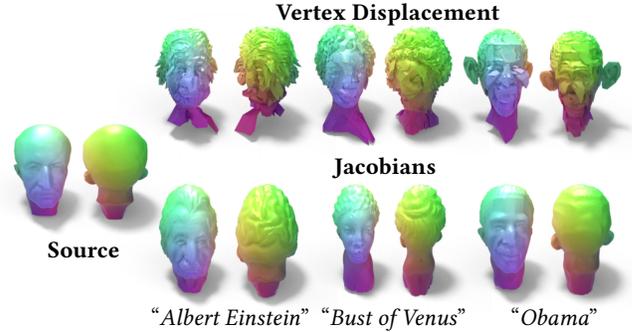

First, straightforward optimization of mesh vertices through differentiable rendering, as previous text-to-3D methods displayed, often converges to undesirable local minima, and gradient steps often turn parts of the mesh inside-out, introducing significant artifacts. The crux of the difficulty lies in that the back-propagated gradient from CLIP is noisy, with many undesirable local minima and arbitrary directions. Thus,
instead of displacing vertices, we take inspiration from Neural Jacobian Fields~\cite{aigerman2022NJF} to devise a more robust representation of deformations. We optimize matrices representing the deformation's gradients, i.e., the Jacobians of each of the triangles, and compute the deformed vertex positions from them, by solving Poisson's equation. Our key observation is that representing the deformation through Jacobians in this way leads to a representation that favors smoother, large deformations, and leads to a global relation between vertices and pixels, thus avoiding localized noisy gradients.
More precisely: i) smooth Jacobians represent low-frequency, large-scale deformations, and ii) Poisson's equation leads to each Jacobian affecting all vertices, and in turn all pixels of the rendering. Thus, when CLIP's gradients back-propagate to the Jacobians, each pixel's gradient has a global effect, leading to a more regularized solution, see \figref{fig:global_jacobian}.

Second, we observe that  CLIP per-pixel embedded features are unfortunately view-dependent and the same 3D coordinate may be assigned significantly different features in different 2D viewpoints. This in turn implies a given vertex on the mesh may receive conflicting gradients from different viewpoints (conceptually, when deforming a cow into an elephant, one viewpoint may try to tug on a vertex to grow a trunk out of it, while the other will try to use it as a tip of a tusk). This leads to incoherent results, and, in some cases, lack of global consistency (e.g., adding multiple trunks when deforming a cow into an elephant).
To counter that, we devise a novel loss, which encourages vertices to achieve similar CLIP features from different viewpoints, thereby leading to global coherency in the deformations.

Third, to ensure the deformed shape still lies in semantic correspondence to the input shape, we add a identity-preserving term, which ensures that the deformation optimization step does not stray too far from the initial input mesh, thereby preventing the optimization from ignoring the input geometry.

Through experiments, we show we can apply text-driven deformations to a large class of source shapes and desired targets (from organic to man-made shapes). \ourmethod{} produces plausible shapes, beyond the capabilities of previous text-driven mesh generation techniques, while additionally providing abilities achieved solely through a deformation framework, such as significant resemblance to the input shape, and providing meaningful correspondences between the source and deformed shapes.

\section{Related Work}
There has been a large variety of works in the space of text-guided content synthesis driven by CLIP~\cite{clip}, a foundational model which learns a joint embedding space for text and images. Many generative models such as StyleCLIP~\cite{styleclip}, GLIDE~\cite{glide}, and DALLE-2~\cite{dalle} leverage text during training by computing distance between text and images in the embedding space of CLIP. Additionally, there has been work on using CLIP guidance in fine-tuning the latest state-of-the-art diffusion models to achieve even higher quality results~\cite{diffusionclip}.

\textbf{Text-Guided 3D Synthesis.}
In comparison to text guided image generation, text-to-3D is relatively undeveloped. While there are some works that propose training joint embeddings of text descriptions and 3D objects~\cite{chen2018text2shape}, these works are lacking in scale as the largest captioned 3D dataset (the recent ObjaVerse dataset with 800k assets~\cite{objaverse}) is  several orders of magnitude smaller than the LAION-5B dataset~\cite{laion}. Nevertheless, there have been large strides in text-to-3D leveraging large pre-trained 2D models such as CLIP.

\begin{figure}[t]
    \centering
    \includegraphics[width=\columnwidth]{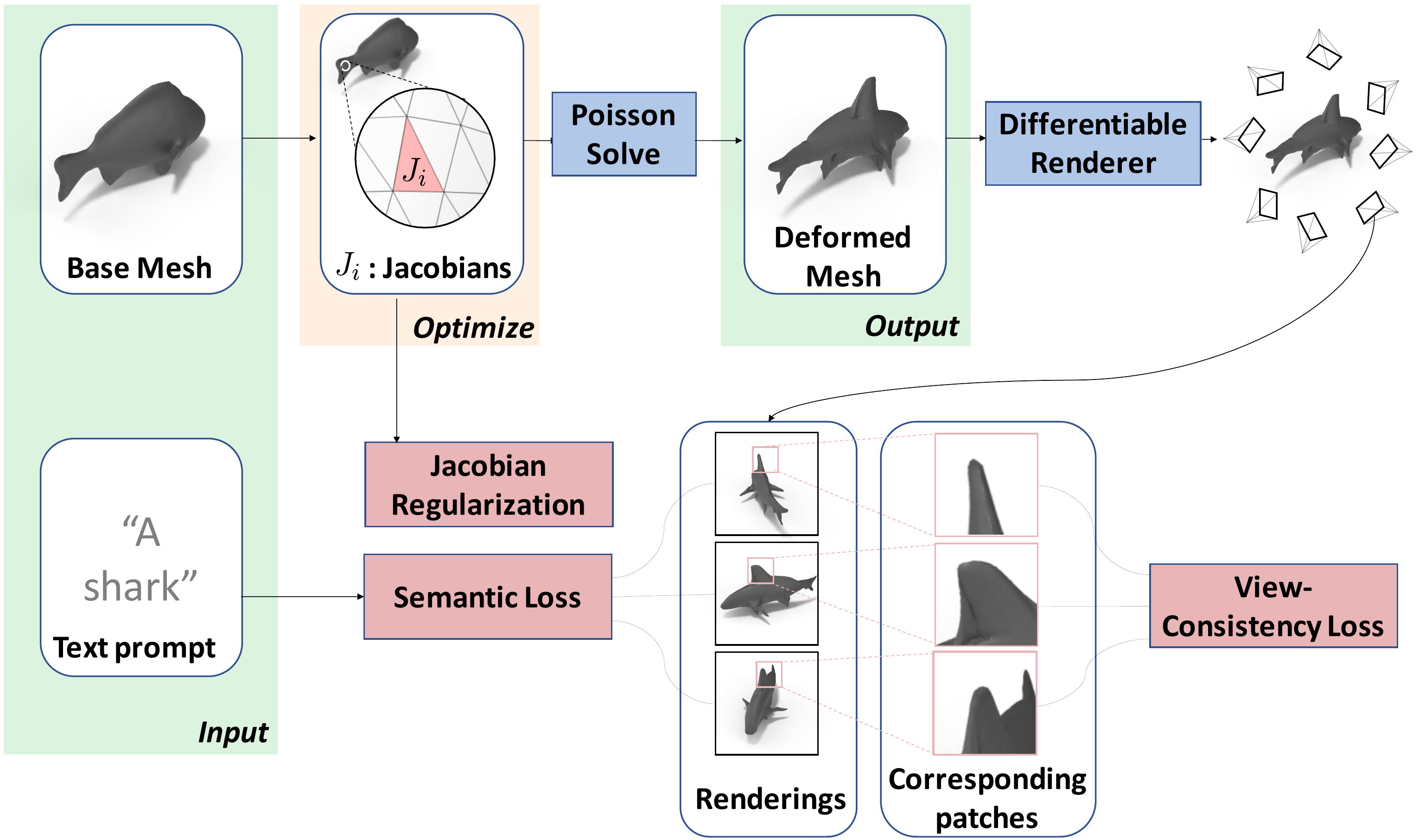}
    \caption{\textbf{Overview.} \ourmethod{} deforms a base mesh by optimizing per-triangle Jacobians using natural language as a guide. We optimize the deformation using three losses: a CLIP-based semantic loss drives the deformation toward the text prompt, a view-consistency loss matches multiple views of the same surface patch to ensure a coherent deformation, and our regularization on the Jacobians controls the fidelity to the base mesh.}
    \label{fig:overview}
\end{figure}
CLIP-Forge~\cite{clipforge} overcomes the lack of pre-trained counterpart to CLIP for 3D by using renderings of training shapes to bridge the gap between text and 3D data. They first train a voxel encoder and an implicit decoder on available 3D datasets using CLIP image embeddings, then swap image embeddings for text embeddings at inference time. However, their method is still limited by the availability of 3D datasets used to train their autoencoder. Point-E~\cite{pointe} proposes to generate an image from text using a 2D diffusion model, then trains a point cloud diffusion model conditioned on images on a private dataset of millions of 3D shapes. While not on par with the state-of-the-art in terms of shape quality, their approach can generate 3D shapes significantly faster.

Several approaches tackle zero-shot geometry synthesis, bypassing the need of a 3D dataset. Dreamfields~\cite{dreamfields} leverage volume rendering with a Neural Radiance Field (NeRF)~\cite{nerf}  to directly optimize views of a 3D shape against a desired text prompt in CLIP's embedding space. Recently, leveraging 2D diffusion models~\cite{stablediffusion, imagen}, DreamFusion~\cite{dreamfusion} and \cite{sjc} propose to distill such models as a differentiable image-based loss. Extracting and editing an explicit mesh from these works is not straightforward, since NeRFs represent shapes through network weights.

Other works have used surface-based differentiable rendering in order to pass views of explicit 3D objects to CLIP. Using this method, Text2Mesh \cite{text2mesh} employs a network to predict colors and deformations along the normals of a template mesh. Their objective is to \textit{stylize} the template mesh while \textit{preserving} the initial content. In contrast, CLIP-Mesh~\cite{clipmesh} proposes deforming the vertices of a sphere in accordance to an input text prompt to synthesize a completely new geometry. Magic3D~\cite{magic3d} combines both representations by first optimizing a radiance field using score-distillation similar to DreamFusion, then extracts an explicit mesh from the radiance field and optimizing its vertices via differentiable surface rendering and score distillation. The code is not public at the time of submission. Our work also leverages differentiable rendering and CLIP, but focuses on the problem of  \textit{deforming} explicit geometry rather than generating it from scratch.

\textbf{Neural Shape Deformation.}
Shape deformation has been traditionally approached  by providing a user with handles that control a deformation space, either through an energy-minimizing formulation~\cite{LaplacianMeshEditing:2004, arap} or through skinning the mesh with weights that interpolate coordinates with respect to the handles~\cite{skinningcourse:2014,Fulton:LSD:2018}. Skinning methods have been used extensively in learning context~\cite{AnimSkelVolNet,RigNet,Holden:inverse_rig:2015,li2021learning},  point handles~\cite{liu2021deepmetahandles}, and cages~\cite{neuralcages}, while variational formulations are used as regularizers, e.g., ARAP~\cite{Sun:2021} or the Laplacian~\cite{cmrKanazawa18}. Both of these approaches span only a subset of possible deformations and prevent fine-grained control over details.

Other methods focus on learning tasks over one template mesh~\cite{tan2018meshvae,gaovcgan2018}, and assign per-vertex coordinates~\cite{Shen:2021} or offsets from a simpler (e.g., linear) model~\cite{Bailey:2018:FDD,bailey2020fast,Romero:2021,Zheng:secondary_motion:2021,yin2021_3DStyleNet}. 

While some recent works propose data-driven approaches to predict realistic deformations~\cite{aigerman2022NJF,jakab2021keypointdeformer, neuralcages, alignet, yumer2015semantic}, the semantic capabilities of all these works are limited once again by the lack of 3D datasets pairing shapes and captions. Our work is similar in spirit to these methods, but instead of requiring explicit supervision, we leverage differentiable rendering and powerful visual models such as CLIP to drive deformations of a template shape.

\section{Method}
\figref{fig:overview} shows an overview of our method. Given an input shape, \ourmethod{}  enables manipulating the geometry guided by a user-specified text description. 

We represent the geometry of the input shape using a mesh $\mathcal{M}$ defined by a set of vertices $\mathcal{V} \in \mathbb{R}^{n\times 3}$ and faces $\mathcal{F}$. We optimize a displacement map $\Phi:\mathbb{R}^3\to\mathbb{R}^3$ over the vertices through differentiable rendering.

\textbf{Deformations through Jacobians.}
A naive optimization of $\Phi$ would simply entail directly displacing each vertex $\mathcal{V}$, which may overly distort the original shape, especially when the target text describes a highly-detailed texture. This is due to this representation exposing high-frequency, oscillatory modes of deformation. Thus, in this work we opt for a different parameterization of deformations. Inspired by Neural Jacobian Fields~\cite{aigerman2022NJF}, we parameterize the shape using a set of per-triangle Jacobians which define a deformation. Specifically, we represent per-triangle jacobians by matrices $J_i\in\mathbb{R}^{3\times 3}$ for every face $f_i\in\mathcal{F}$. Following~\cite{aigerman2022NJF}, we solve a Poisson problem to compute a deformation map $\Phi^\ast$ as the mapping with Jacobian matrices for each face that are closest to $\{J_i\}$ in the least-squares sense, that is:
\be
    \Phi^\ast = \min_{\Phi} \sum_{f_i\in\mathcal{F}}|f_i|\lVert \nabla_i(\Phi) - J_i\rVert_2^2
    \label{eq:poisson_equation}
\ee
where $\nabla_i(\Phi)$ denotes the Jacobian of $\Phi$ at triangle $f_i$ and $|f_i|$ is the area of that triangle. Hence, we may optimize the deformation mapping $\Phi$ indirectly by optimizing the matrices $\{J_i\}$ which define $\Phi^\ast$. These Jacobians are initialized to identity matrices, thereby initializing  $\Phi^\ast$ as the identity mapping. Please refer to~\cite{aigerman2022NJF} for the full technical details.

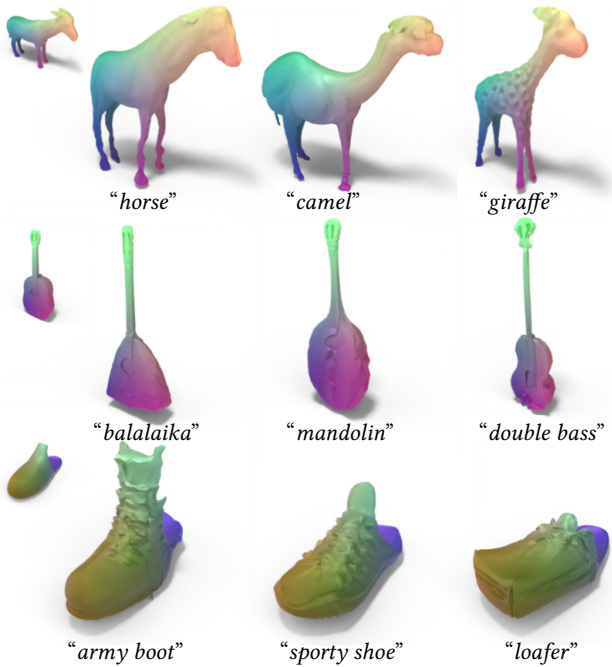
\begin{figure}[t]
    \centering
    \newcommand{\pl}{-1.5}
    \begin{overpic}[width=\columnwidth]{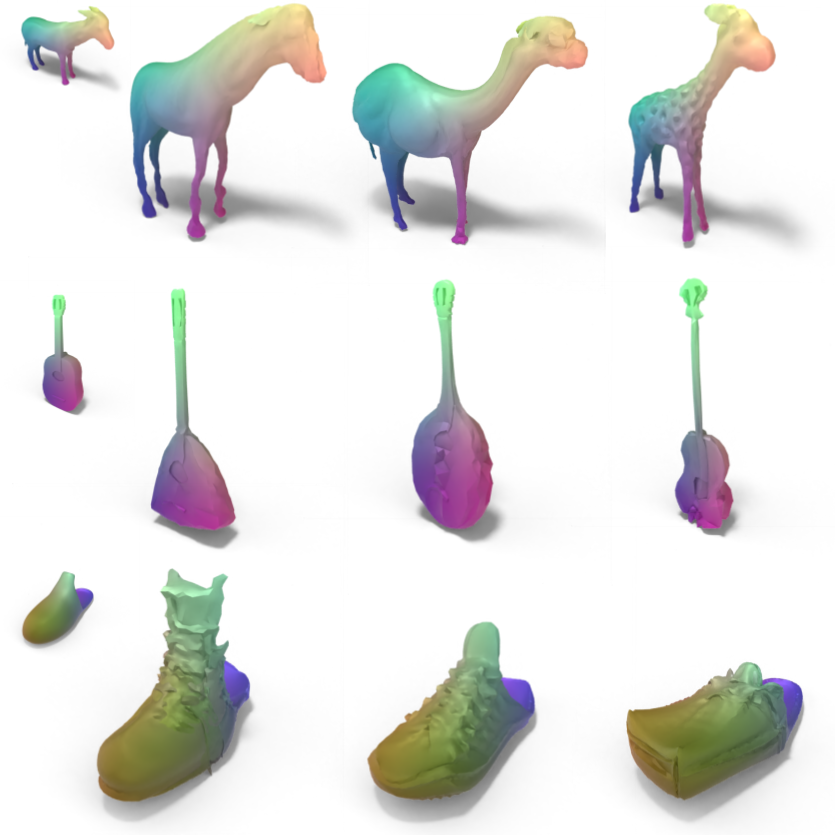}
    \put(18.5,  68){\textcolor{black}{``\textit{horse}''}}
    \put(46,  68){\textcolor{black}{``\textit{camel}''}}
    \put(76,  68){\textcolor{black}{``\textit{giraffe}''}}
    \put(16,  32){\textcolor{black}{``\textit{balalaika}''}}
    \put(46,  32){\textcolor{black}{``\textit{mandolin}''}}
    \put(76,  32){\textcolor{black}{``\textit{double bass}''}}
    \put(12, -2){\textcolor{black}{``\textit{army boot}''}}
    \put(45, -2){\textcolor{black}{``\textit{sporty shoe}''}}
    \put(80, -2){\textcolor{black}{``\textit{loafer}''}}
    \end{overpic}
    \vspace{-1pt}
    \caption{\textbf{Text-driven deformation.} Our method can deform the same source into different text-specified geometries, achieving local geometric details (shoe laces and giraffe) and low-frequency shape modifying deformations (body of guitar, shape of animal).}
    \label{fig:same_source}
\end{figure}

\textbf{Language Guidance.}
Our objective is to use text to guide the deformation of the source shape. We leverage the pre-trained vision-language CLIP~\cite{clip}, which provides a shared embedding space between images and text. In order to connect geometry to images, we pass our shape through a differentiable renderer $\mathcal{R}$~\cite{nvdiffrast}. Hence, we may differentiably embed the renders of the deformed shape $$e_{\mathcal{M}} = \texttt{CLIP}(\Phi^\ast(\mathcal{M})) \in \mathbb{R}^{512}$$ We abuse notation and the rendering function $\mathcal{R}$ is understood to be implicit when passing shapes to $\texttt{CLIP}$. The desired deformation, described with a natural language prompt $\mathcal{P}$, is also embedded $e_\mathcal{P} = \texttt{CLIP}(\mathcal{P}) \in \mathbb{R}^{512}$. Then, we may optimize $\Phi^\ast$ such that $e_\mathcal{M}$ and $e_\mathcal{P}$ agree, by maximizing the cosine similarity between the embeddings:
\be
    \mathcal{L}_{\mathcal{P}}(\Phi^\ast, \mathcal{M}, \mathcal{P}) = \mathrm{sim}\left(e_{\mathcal{M}}, e_{\mathcal{P}}\right)
    \label{eq:identity}
\ee
where $\text{sim}(\cdot,\cdot)$ stands for cosine similarity. 
Similarly to  StyleCLIP~\cite{styleclip}, we %
find that incorporating relative directions in CLIP's embedding space can give stronger signals when the optimization landscape between $\Phi^\ast$ and $\mathcal{P}$ is unclear. Given a base caption $\mathcal{P}_0$ that describes $\mathcal{M}$, we compute the direction between the target prompt and the base prompt: $\Delta\text{CLIP}(\mathcal{P}, \mathcal{P}_0) = \texttt{CLIP}(\mathcal{P}) - \texttt{CLIP}(\mathcal{P}_0)$. We  compute the direction of the deformations and aim to optimize: 
\be
    \mathcal{L}_{\Delta\mathcal{P}}(\Phi^\ast, \mathcal{P}, \mathcal{P}_0) = \text{sim} \left(\Delta\text{CLIP}(\mathcal{P}, \mathcal{P}_0), \Delta\texttt{CLIP}(\Phi^\ast(\mathcal{M}), \mathcal{M})\right).
\ee

\begin{figure}[t]
    \centering
    \includegraphics[width=\columnwidth]{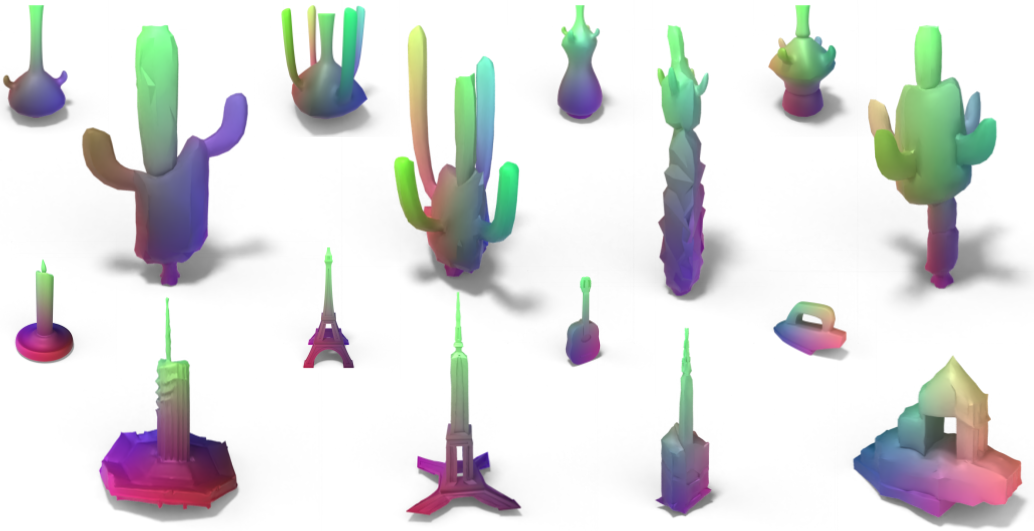}
    \caption{Deformations of different source vases into the same target text, ``\textit{a cactus}'' and different source meshes into the same target text, ``\textit{a skyscraper}''. Our method meaningfully respects the input geometry while conforming to the desired target text. }
    \label{fig:same_target}
\end{figure}
\textbf{Jacobian Regularization.}
To prevent the deformation from straying too far from the input undeformed geometry, we introduce another regularization term on the predicted Jacobians, which penalizes the difference between the Jacobians $\{J_i\}$ and the identity, i.e., no deformation:
\be
    \mathcal{L}_{I}(t_j) = \alpha\sum_{i=1}^{|\mathcal{F}|} \lVert J_i - I\rVert_2
\ee
where $\alpha$ is a hyper-parameter which may be tuned to control the strength of the deformations defined by $\{J_i\}$.

\textbf{View Consistency.} A common problem when performing multi-view optimization through CLIP is the lack of view-consistency, \textit{i.e.}, a particular view may pull the shape toward a specific deformation while another view pulls toward a different deformation. Averaging gradients on the Jacobians for each view does not necessarily lead to a coherent 3D deformation which may manifest in various artifacts, including muddled details and incorrect geometry. 

\begin{figure*}[t]
    \centering
    \begin{overpic}[width=\textwidth]{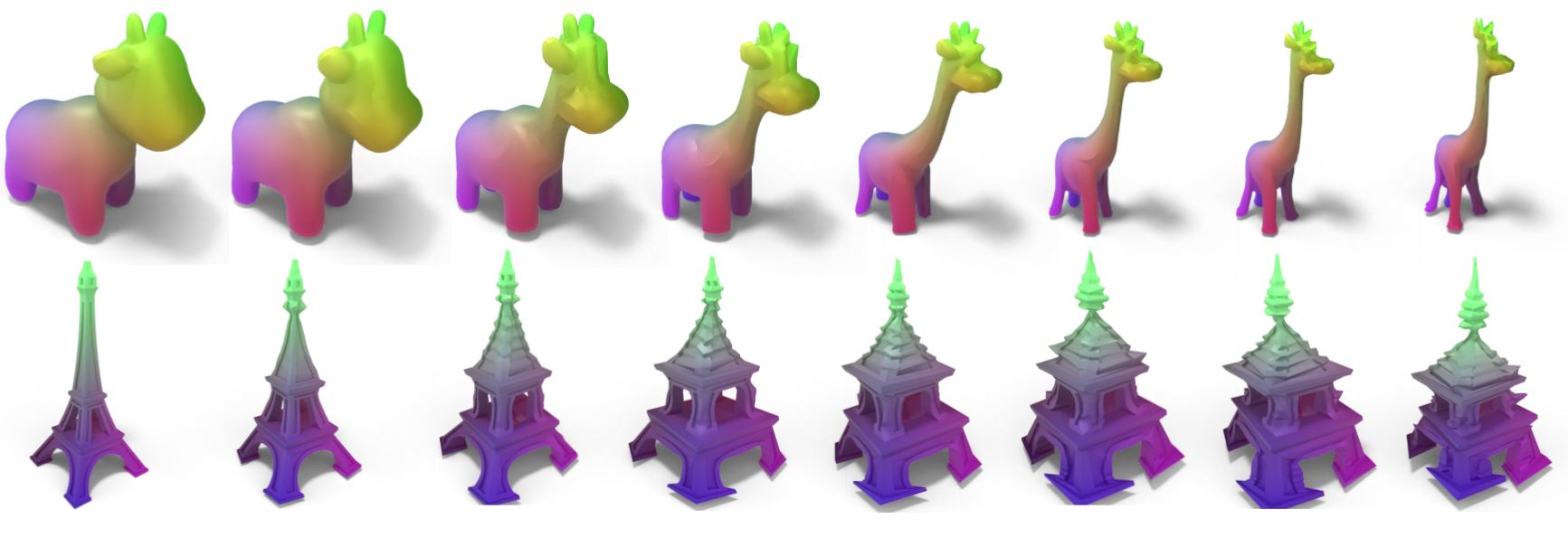}
    \put(3, 0){\textcolor{black}{\textbf{Source}}}
    \put(16,0){\textcolor{black}{$\alpha=25$}}
    \put(30,0){\textcolor{black}{$\alpha=10$}}
    \put(43,0){\textcolor{black}{$\alpha=5$}}
    \put(55.5,0){\textcolor{black}{$\alpha=2$}}
    \put(68.5,0){\textcolor{black}{$\alpha=1$}}
    \put(79.5,0){\textcolor{black}{$\alpha=0.5$}}
    \put(92.5, 0){\textcolor{black}{$\alpha=0$}}
    \end{overpic}
    \caption{\textbf{Identity preservation.} The source shape (left) is deformed into a ''\textit{giraffe}'' (top), ''\textit{pagoda}'' (bottom) using decreasing amount of weight on the proposed Jacobian regularization. Higher weight encourages preservation of the identity of the source. Note that results with zero Jacobian regularization (right column) may contain artifacts.}
    \label{fig:arap_stress_test}
\end{figure*}
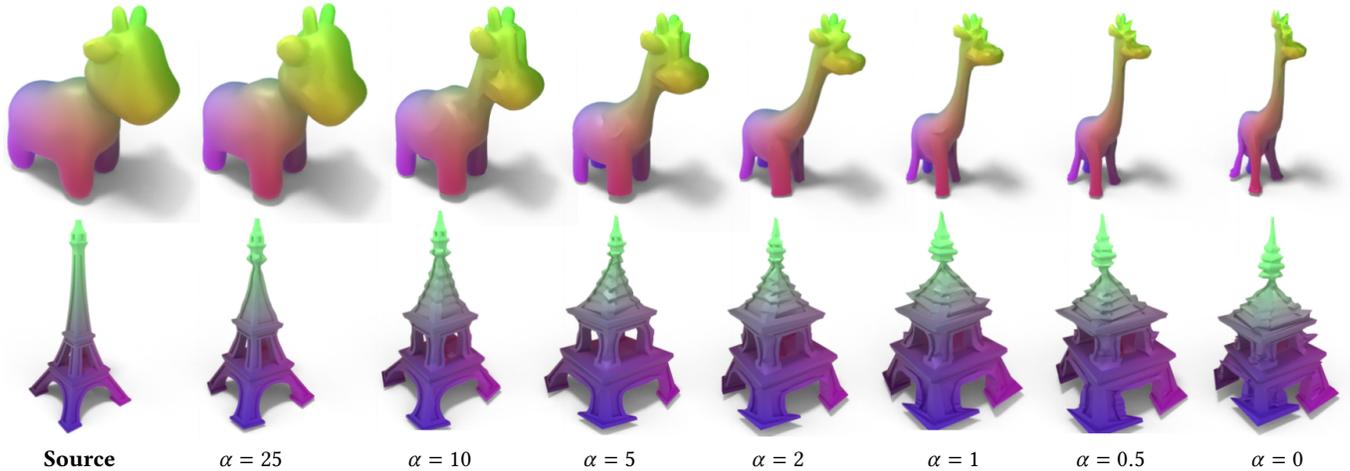

We introduce another regularization term to tackle this problem by utilizing the patch-level deep features of CLIP's vision transformer (ViT). In ViTs, the image is split into non-overlapping patches $P_0, P_1, \dots P_n$, which are then projected into a higher-dimensional space and passed through transformer encoder blocks $\mathcal{T}_1, \mathcal{T}_2, \dots \mathcal{T}_n$. For each vertex $v\in\mathcal{V}$, and each render $r\in \mathcal{R}(\mathcal{M})$, if $v$ is visible, we compute the pixel $p(v, r)$ in $r$ that contains $v$. Then, by associating $p(v, r)$ with the nearest corresponding patch center $P(v, r)$, we extract a deep feature vector corresponding to $v$ and $r$ %
and encourage vertices to have similar deep features across renders from different viewpoints:
\be
\mathcal{L}_{\text{VC}}(v) = \sum_{i=1}^{|\mathcal{R}(\mathcal{M})|} \sum_{\substack{j=1\\j\neq i}}^{|\mathcal{R}(\mathcal{M})|} \text{sim} \left(\mathcal{T}_k(P(v, r_i)), \mathcal{T}_k(P(v, r_j))\right)
\ee
for some chosen layer $\mathcal{T}_k$. In practice, we choose to use the \textit{token} output of the final transformer block.  Then, we simply penalize this loss over all vertices $v \in\mathcal{M}$:
\be
\mathcal{L}_{\text{VC}}(\mathcal{M}) = \beta\sum_{v\in\mathcal{V}}\mathcal{L}_{\text{VC}}(v)
\ee

where $\beta$ is another tunable hyper-parameter. To compute $P(v, r)$, we follow~\cite{deepvitfeatures} by modifying CLIP's ViT to use a smaller stride in its initial convolution, obtaining \textit{overlapping} patches. Then, by interpolating the positional encoding, we achieve finer-resolution deep features. %

\section{Experiments}
We run \ourmethod{} on a variety of text prompt, source mesh pairs. Each pair takes approximately 1.5 hours for 5000 iterations.
\label{sec:exp}
\subsection{Generality of \ourmethod{}}
We show that \ourmethod{} can handle a wide variety of source and target prompts. We represent the source mesh as a small inset next to the deformed shapes unless otherwise specified.~\figref{fig:gallery} shows one such collection of source and targets. We see that \ourmethod{} is capable of producing high-quality results for different types of source and target pairs.

\textbf{Adjective Targets.} We see that \ourmethod{} is capable of deforming source meshes in accordance to a target \textit{adjective}. In~\figref{fig:gallery}, we see that \ourmethod{} deforms a cat to be ''\textit{fluffy},'' a vase to be ``\textit{heart-shaped},'' an alien to be ``\textit{scary},'' and a shoe to be ``\textit{high-heeled}.'' For this style of target text, our method produces deformation maps that preserve the overall structure of the source shape.

\textbf{Related Targets.} \ourmethod{} is also capable of deforming source meshes into \textit{related} shapes, which are not exactly descriptors of the original mesh, but also not too semantically different. For example, in~\figref{fig:gallery}, we see that \ourmethod{} is able to deform an acoustic guitar into a ``\textit{banjo},'' a candle into an ``\textit{incense stick},'' the Eiffel tower into a ``\textit{pagoda},'' a vase into a ``\textit{royal goblet},'' and so on.

\textbf{Unrelated Targets.} Finally, we observe that \ourmethod{} is capable of deforming source meshes into completely \textit{unrelated} shapes, which are far from the source mesh. This capability is illustrated in~\figref{fig:gallery} where an ant is deformed into a ``\textit{snake},'' and a clothing iron is deformed into a ``\textit{kettle},'' as well as in~\figref{fig:spot} where a cow is deformed into various unrelated animals such as a ``\textit{turtle}''.

\begin{figure}[b]
    \centering
    \includegraphics[width=\columnwidth]{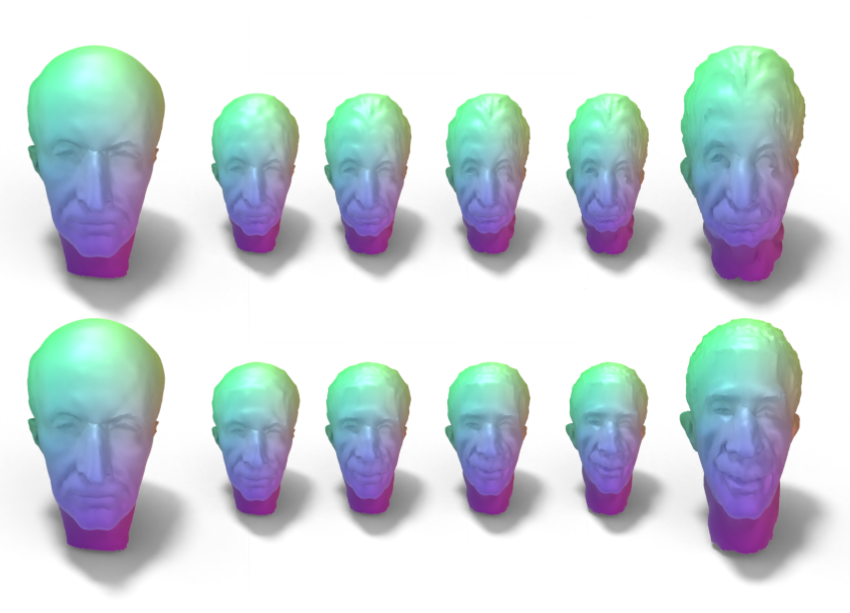}
    \caption{Visualizing iterations of the optimization as the source mesh is deformed to the target (``\textit{Einstein}'', ``\textit{Obama}''). Due to our formulation and energy, each facial feature of Max Planck is deformed into the corresponding facial feature of the target (nose to nose, eyes to eyes etc.).}
    \label{fig:my_label}
\end{figure}
\subsection{Expressiveness of \ourmethod{}}
\begin{figure*}[t]
    \centering
    \begin{overpic}[width=\textwidth]{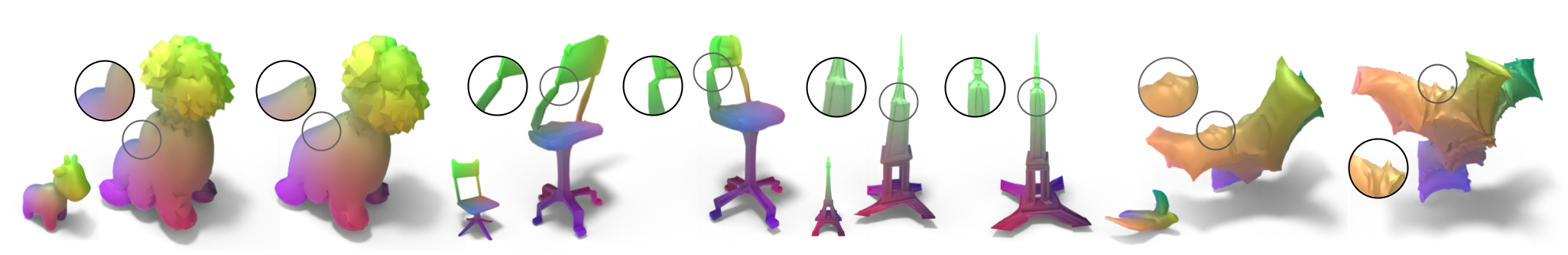}
    \put(9.5, -0.5){\textcolor{black}{``\textit{fluffy poodle}''}}
    \put(7.5, 15.5){\textcolor{black}{No $\mathcal{L}_{\text{VC}}$}}
    \put(19, 15.5){\textcolor{black}{With $\mathcal{L}_{\text{VC}}$}}
    \put(36.5, -0.5){\textcolor{black}{``\textit{gaming chair}''}}
    \put(33, 15.5){\textcolor{black}{No $\mathcal{L}_{\text{VC}}$}}
    \put(42, 15.5){\textcolor{black}{With $\mathcal{L}_{\text{VC}}$}}
    \put(57, -0.5){\textcolor{black}{``\textit{skyscraper}''}}
    \put(54, 15.5){\textcolor{black}{No $\mathcal{L}_{\text{VC}}$}}
    \put(62, 15.5){\textcolor{black}{With $\mathcal{L}_{\text{VC}}$}}
    \put(84, -0.5){\textcolor{black}{``\textit{bat}''}}
    \put(75.5, 15.5){\textcolor{black}{No $\mathcal{L}_{\text{VC}}$}}
    \put(88, 15.5){\textcolor{black}{With $\mathcal{L}_{\text{VC}}$}}
    \end{overpic}
    \caption{\textbf{Viewpoint Consistency Ablation.} Ablation results of removing $\mathcal{L}_\text{VC}$ for four different source-text pairs. In each example we see instances of incorrect geometry in the shapes deformed without $\mathcal{L}_\text{VC}$. }
    \label{fig:vp_ablation}
\end{figure*}
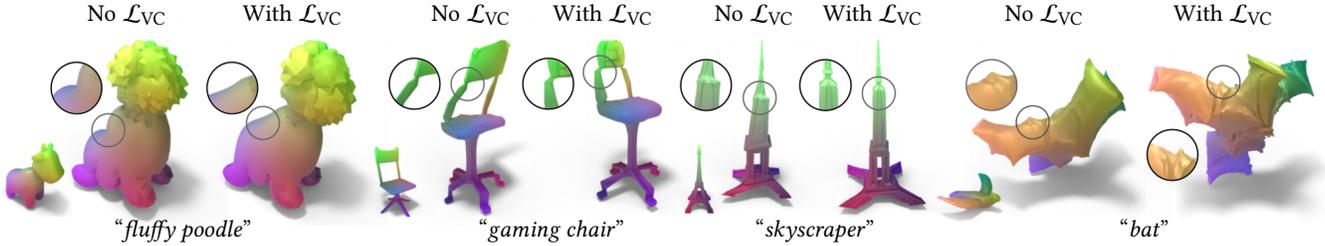
\textbf{Frequency.} We study the expressiveness of our method by experimenting with different text prompts for the same source mesh. We observe that \ourmethod{} is powerful enough to express both high-frequency texture details as well as low-frequency deformations to the shape structure. In~\figref{fig:same_source}, we see in the top row that \ourmethod{} can produce the requisite low-frequency deformations to change the donkey into various other animal shapes, but also  add high-frequency deformations to emulate ``\textit{giraffe}'' spots. Similarly in middle row, \ourmethod{} produces different low-frequency maps to change the shape of the guitar body while preserving the neck. Alternately in the last row, \ourmethod{} produces high-frequency deformations to emulate fine details such as the laces of an ``\textit{army boot}'' or the creases in a ``\textit{sporty shoe}.'' Similarly, we observe \ourmethod{} producing high-frequency deformations in~\figref{fig:spot}, particular in the ``\textit{alligator}'' example.

\textbf{Dense Matching.} In~\figref{fig:my_label}, we demonstrate the ability of \ourmethod{} to preserve highly detailed semantics throughout the optimization process. Not does \ourmethod{} correctly deform features of Max Planck to the corresponding features of the target (``\textit{Einstein},'' ``\textit{Obama}''), it is consistent at every step. We observe at different intermediate faces that the corresponding features are always mapped correctly.

\subsection{Identity Preservation}

We demonstrate our method's ability to preserve the source shape in two ways. 

\textbf{Impact of the Input Geometry.} In~\figref{fig:same_target}, we use \ourmethod{} to deform different vases with various appendages into ``\textit{a cactus}.'' Note that the resulting deformation map grows cactus branches depending on where appendages are on the source mesh. Hence, \ourmethod{} is able to produce a deformation map that conforms to the target text, ``\textit{cactus},'' in an adaptive manner, borrowing coarse structures from the input geometry. We also observe a similar effect when deforming semantically diverse meshes into the target text, ``\textit{skyscraper}''. \ourmethod{} preserves different aspects of each mesh, such as the base of the candle and the body of the guitar. It also produces interesting variation \eg deforming the flatter iron into a dome structure instead of a thinner needle.

\textbf{Jacobian Regularization.} Recall that we also define a Jacobian regularization term $\mathcal{L}_I$ in~\eqref{eq:identity} which is scaled by a hyper-parameter $\alpha$. In~\figref{fig:arap_stress_test}, we show that adjusting $\alpha$ controls how far the deformation map $\Phi^\ast$ is allowed to deviate from the source mesh. With $\alpha=25$, we see that the cow and the Eiffel tower do not change meaningfully in accordance to their respective text prompts (``\textit{giraffe}'' and ``\textit{pagoda}''), while setting $\alpha=0$ may result in some artifacts in the deformed shape. We observe that setting $\alpha$ to intermediate values offers the best results.

\subsection{Viewpoint Consistency}
\label{sec:vp}

We experiment with the qualitative effect of the viewpoint consistency loss $\mathcal{L}_{\text{VC}}$ by using \ourmethod{} with and without this loss term and noting the differences in the deformed meshes. We observe in Figure~\ref{fig:vp_ablation} that the deformations produced without $\mathcal{L}_\text{VC}$, although smooth due to our choice in representation, often contain unrealistic geometric features. Such abnormalities can be caused by outliers in the sampled camera views during optimization. This problem is especially apparent in the ``\textit{gaming chair},'' in which the backrest is crooked. This inaccuracy may appear correct in some perspectives, but is overall an undesirable feature. Another example is the tip of the ``\textit{skyscraper},'' which is pulled to one side of the building during optimization. Finally, we observe that the back of the ``\textit{fluffy poodle},'' and the wings of the ``\textit{bat}'' are incorrectly curved inwards when they are optimized without $\mathcal{L}_\text{VC}$. We provide a quantitative evaluation of these observations in~\secref{sec:quant}.

\begin{figure}[b]
    \centering
    \begin{overpic}[width=\columnwidth]{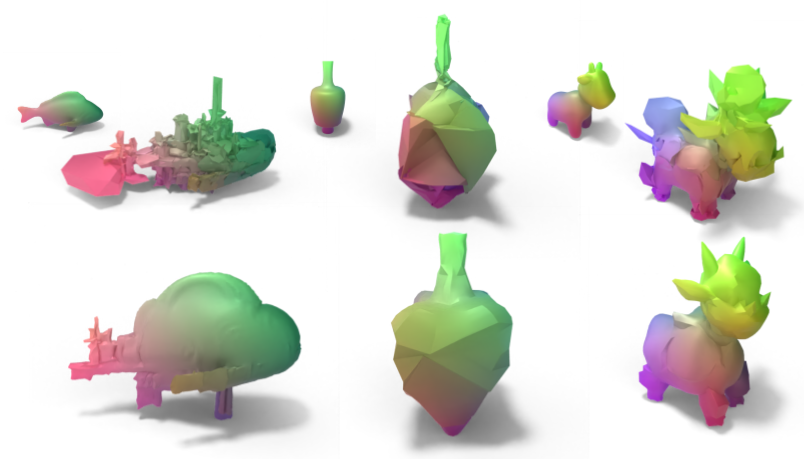}
    \put(17, -0.5){\textcolor{black}{``\textit{submarine}''}}
    \put(40, -0.5){\textcolor{black}{``\textit{diamond-shaped vase}''}}
    \put(76, -0.5){\textcolor{black}{``\textit{cow Pokémon}''}}
    \end{overpic}
    \caption{\textbf{Deformation Ablation.} Top row: \ourmethod{} with vertex displacements. Bottom row: \ourmethod{} with Jacobians. Jacobians are crucial to preserving the structure of the input geometry and maintaining high-quality surfaces.}
    \label{fig:vertex_deformation}
\end{figure}

\subsection{Effect of Jacobians} %
We also experiment with replacing Jacobians with vertex displacements in our pipeline, but keeping the additional losses we introduce, $\mathcal{L}_{\Delta \mathcal{P}}$ and $\mathcal{L}_\text{VC}$.

\begin{figure}[h]
    \centering
    \begin{overpic}[width=\columnwidth]{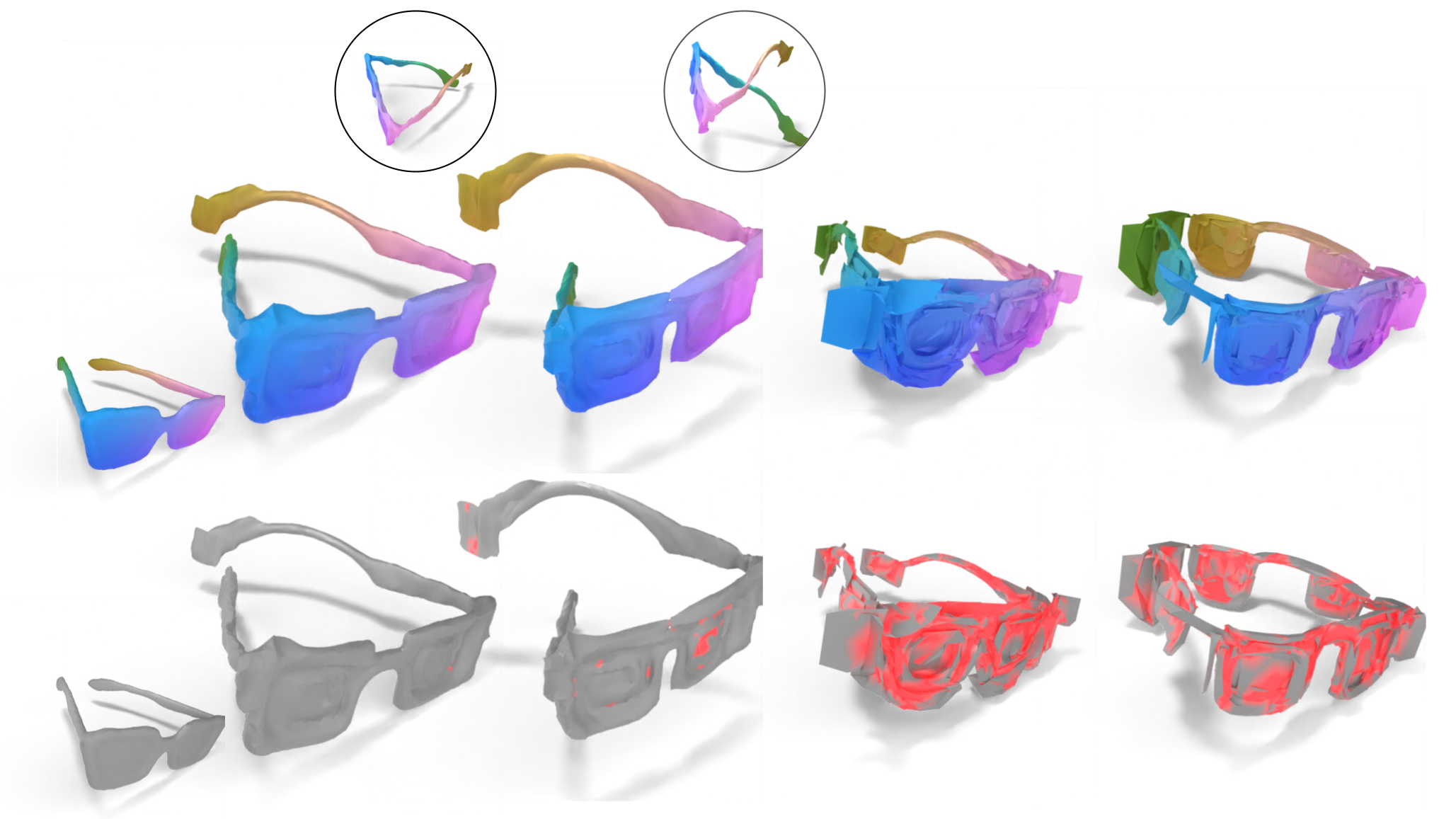}
    \put(23,0){\textcolor{black}{Ours}}
    \put(40,0){\textcolor{black}{No VC}}
    \put(54,0){\textcolor{black}{Verts (with VC)}}
    \put(80,0){\textcolor{black}{CLIP-Mesh}}
    \end{overpic}
    \caption{\textbf{Self-intersections.} Comparison results for the shown source and target text ``goggles". Self-intersections are highlighted in red (bottom row). Removing view-consistency (VC) losses causes distortion on the temple arms. Removing Jacobians and optimizing vertices introduces further surface distortion and self-intersections which may impede utility. When applying CLIP-Mesh to this template, we observe the ``janus'' effect e.g. unrealistic repeated geometry on each side.}
    \label{fig:mesh_quality}
\end{figure}
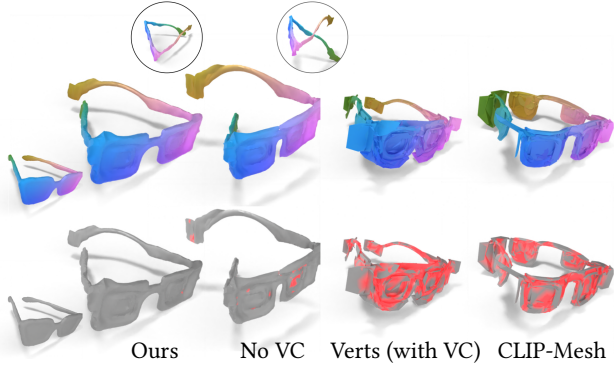

\textbf{Surface Quality.} We observe that our use of Jacobians plays a key in obtaining high-quality surface geometries.  In \figref{fig:vertex_deformation}, we show that replacing Jacobians with vertex displacements in \ourmethod{} drastically decreases the resulting surface quality for three different examples.  In the ``\textit{diamond-shaped vase},'' vertex displacements cause the shape to collapse inwards. In the other two examples, vertex displacements create details which do not respect the initial shape template and contain many self-intersections. The bottom row of~\figref{fig:mesh_quality} highlights this deterioration of surface quality for another example. We observe that vertex displacements cause a large number of self-intersections, even in areas where the resulting surface appears to require a relatively simple deformation, such as the lens of the ``\textit{goggles}.''

\textbf{Globally-Coherent Deformations.} Recall from~\eqref{eq:poisson_equation} that we solve a global Poisson system to compute the deformation map from the Jacobians. Hence, unlike the gradients of vertices which only affect one point, gradients propagated through Jacobians influence a large surface mesh area, leading to a \textit{globally-coherent} deformation. In~\figref{fig:global_jacobian}, in which a bust of Max Planck is deformed to various other faces, we observe that vertex displacements take local, sub-optimal steps, not only contributing to poor surface quality, but also, in the example of ``\textit{Obama},'' causing spurious face mirroring on the back of the head. Since there are few camera views in which the front and back of the head are visible, the optimization process naturally leads to this result. However, we observe that Jacobians do not experience this artifact. 

\begin{table}[h]
\vspace{-0.1cm}
\newcommand{\allcolor}{\color[rgb]{0.4,0.4,0.95}}
\centering
\begin{tabular}{lccc} 
\toprule
& CLIP R-Precision (L/14) $\uparrow$ & Intersections $\downarrow$  \\
\toprule
Ours & 55.2\% &  3.2\% \\
Ours-noVP & 51.5\% &  3.3\% \\
Ours-Verts & 55.4\% &  67.7\% \\
CLIP-Mesh & 57.4\% &  62.8\% \\
Text2Mesh & 12.7\% & 17.3\%\\
\bottomrule
\end{tabular}
    \caption{\textbf{Quantitative evaluation.} We use our text prompts and deformed meshes in a retrieval task to compute R-Precision. We observe that regularizing for viewpoint consistency improves \ourmethod{} R-Precision. \ourmethod{} and CLIP-Mesh achieve quantitatively comparable R-Precision, \ourmethod{} (Ours) produces higher-quality geometry both qualitatively (see \figref{fig:mesh_quality}) and quantitatively (significantly fewer self-intersections). All methods significantly outperform Text2Mesh in R-Precision.}
\label{tab:quant}
\vspace{-0.3cm}
\end{table}
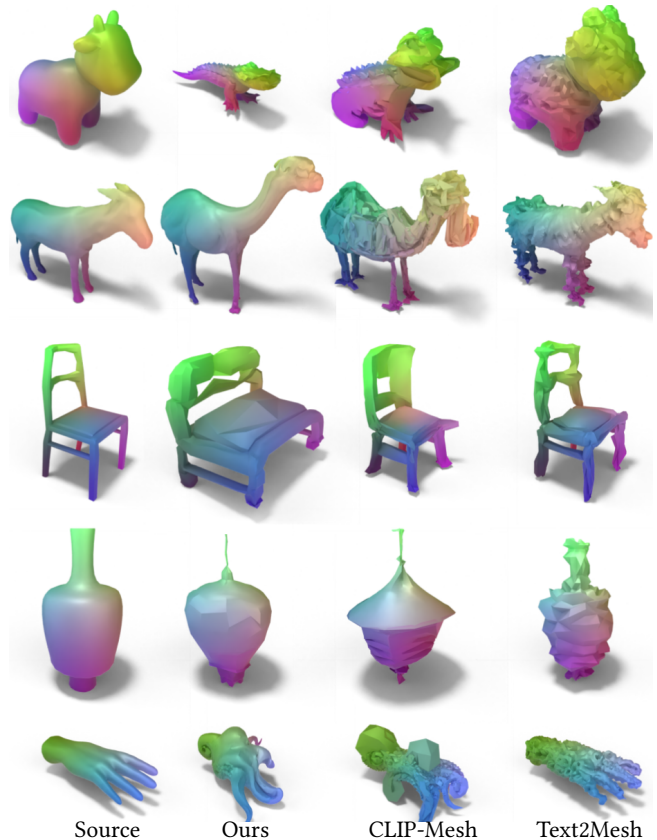
\begin{figure}[h]
\begin{overpic}[width=\columnwidth]{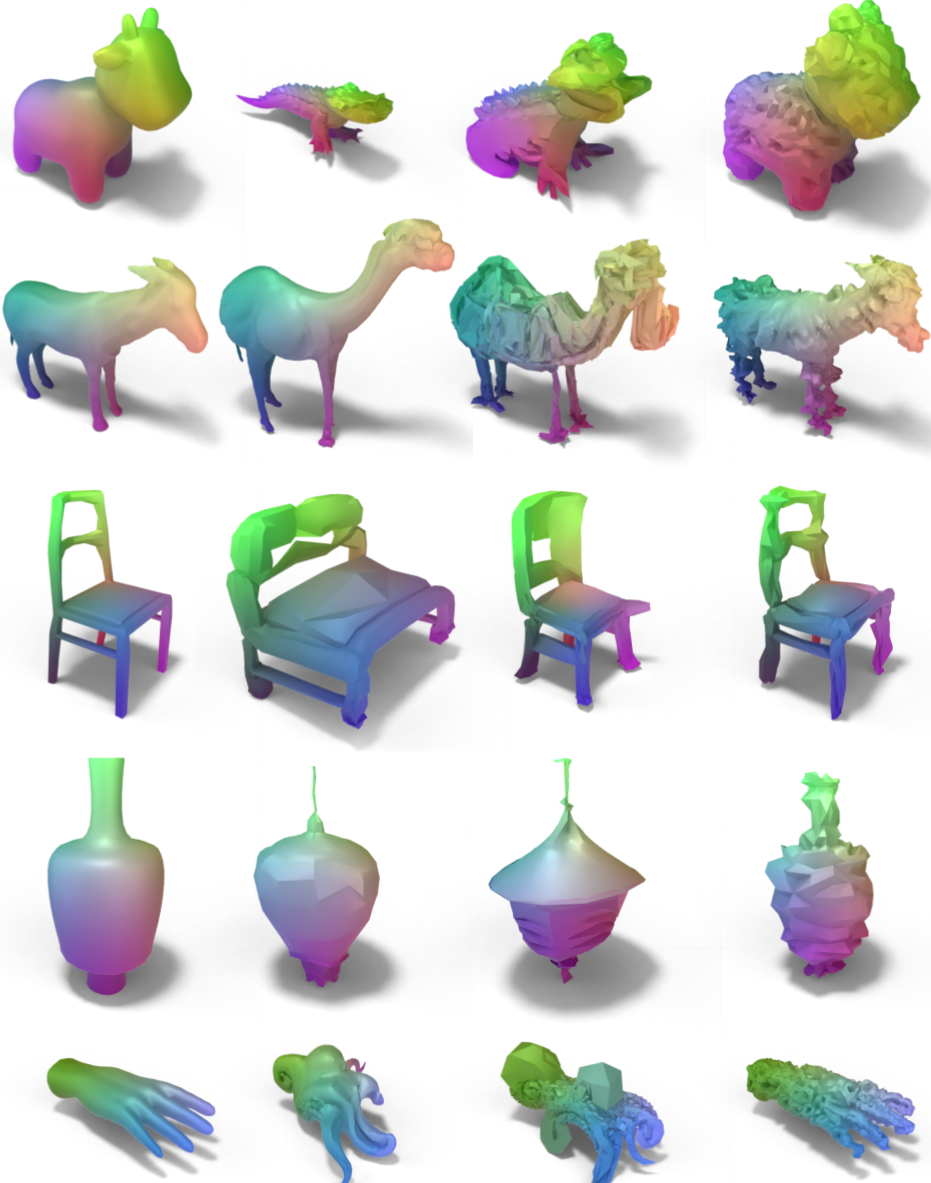}
\put(8, -1.8){\textcolor{black}{Source}}
\put(26, -1.8){\textcolor{black}{Ours}}
\put(44, -1.8){\textcolor{black}{CLIP-Mesh}}
\put(64.5, -1.8){\textcolor{black}{Text2Mesh}}
\end{overpic}
\caption{\textbf{Qualitative Comparison} From top to bottom, the target text prompts are ``\textit{an alligator}'', ``\textit{a camel}'', ``\textit{a comfortable chair}'', ``\textit{a chinese lantern}'', ``\textit{an octupus}''. Compared to CLIP-Mesh, our method produces more semantically correct and higher quality surfaces. Text2Mesh fails to produce semantically meaningful deformations in all examples.}
\label{fig:qual_comp}
\end{figure}

\subsection{Qualitative Comparisons}\label{sec:qual}
Beyond ablations, we also qualitatively evaluate our method against two existing  text-based methods for 3D synthesis and editing, CLIP-Mesh and Text2Mesh. In~\figref{fig:qual_comp}, we show results for 5 different target text prompts: ``\textit{an alligator}'', ``\textit{a camel}'', ``\textit{a chinese lantern}'', ``\textit{a comfortable chair}'', and ``\textit{an octupus}''. \ourmethod{} outperforms the baselines in multiple aspects. First, our method produces more semantically accurate deformations: a more elongated ``\textit{alligator}'', a larger ``\textit{comfortable chair}'', and a rounder ``\textit{chinese lantern}''. In the examples of ``\textit{a camel}'' and ``\textit{an octopus}'' where CLIP-Mesh produces semantically correct results, we observe that it produces poor quality surfaces with irregular triangulation and self-intersections, whereas \ourmethod{} produces smoother, more realistic geometry.

\subsection{Quantitative Evaluation}\label{sec:quant}
We evaluate \ourmethod{} quantitatively in two ways, comparing to CLIP-Mesh~\cite{clipmesh}, Text2Mesh~\cite{text2mesh}, as well as our method without $\mathcal{L}_\text{VC}$ and our method but using vertex displacements.

\textbf{Retrieval Precision.} First, following CLIP-Mesh~\cite{clipmesh}, we compute the R-Precision of CLIP-L/14 on a retrieval task. Specifically, we use a set of 111 text prompt, source mesh pairs to generate a set of deformed meshes. Then, we retrieve deformed shapes for each text prompt through CLIP (L/14) cosine similarity. The first column of~\tabref{tab:quant} shows the results of this experiment. We observe that the viewpoint consistency loss $\mathcal{L}_\text{VC}$ increases the deformation quality of the model, affirming our observations in~\secref{sec:vp}. We also observe that the pipelines using vertex displacements achieve the highest R-Precision scores. This is to be expected as vertex displacements give a high amount of freedom when deforming the shape towards the text prompt, without respect to the initial template geometry. Finally, all methods outperform Text2Mesh, which is more suited for stylization and texturization problems rather than our semantic editing problem.

\textbf{Geometric Quality (Self-Intersections).} We also measure the ratio between the number of self-intersections in the deformed mesh and the number of faces. This validates the qualitative evaluation of in~\figref{fig:vertex_deformation},~\figref{fig:qual_comp}, and~\figref{fig:mesh_quality} that the vertex displacement pipelines disregard the triangulation of the shape in order to optimize CLIP cosine-similarity.

\subsection{Qualitative Comparison with Stable Dreamfusion}\label{sec:qual_sd}

\begin{figure}[t]
    \centering
    \begin{overpic}[width=\columnwidth]{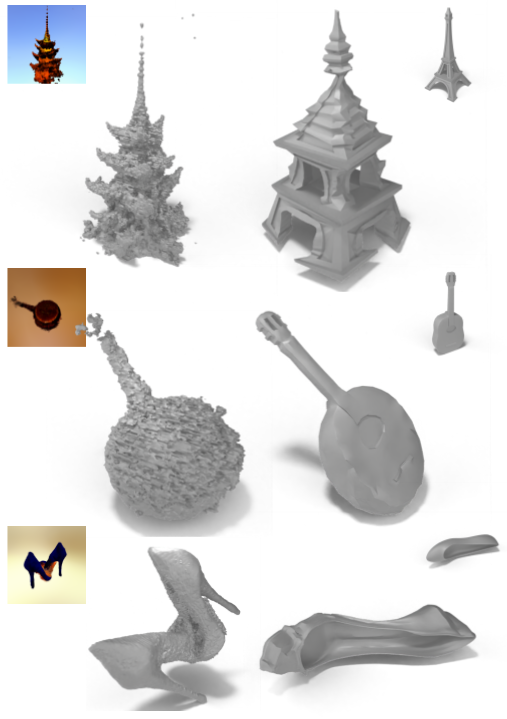}
    \put(0, 10){\textcolor{black}{``\textit{high heel}''}}
    \put(0, 40){\textcolor{black}{``\textit{banjo}''}}
    \put(0, 75){\textcolor{black}{``\textit{pagoda}''}}
    \end{overpic}
    \caption{\textbf{Comparison to Stable Dreamfusion} We compare \ourmethod{} against the open-source implementation of Dreamfusion~\cite{stable-dreamfusion} based on Stable Diffusion~\cite{stablediffusion}. For Stable Dreamfusion, we show  a representative neural rendering as well as the extracted geometry. For \ourmethod{}, we show the initial mesh. First, notice that the surface of Dreamfusion meshes has heavy artifacts compared to the smoothness of our deformations obtained through Jacobians. Second, we notice that Dreamfusion suffers frequently from the Janus problem (see the high heels for instance) which we help alleviate with our View-Consistency loss.
    }
    \label{fig:stabledreamfusion}
\end{figure}

In Figure~\ref{fig:stabledreamfusion}, we compare qualitatively against Stable Dreamfusion. We use the third-party open source implementation of the method~\footnote{https://github.com/ashawkey/stable-dreamfusion}~\cite{stable-dreamfusion}.
We show the geometry extracted from the neural radiance field with the representative neural image as an inset. We observe that the surfaces of geometries from Dreamfusion lack smoothness, which we naturally obtain through our Jacobian representation. Second, we observe that DreamFusion often suffers from the Janus effect, even wiht the use of prompt augmentation on the renderings, whereas our View-Consistency loss helps produce more coherent geometry.

\section{Discussion and Future Work}
In this paper, we propose \ourmethod{}, a zero-shot text-driven mesh deformation technique which does not need to be trained on \textit{any} 3D dataset or 3D annotations. Instead, it is guided by pre-trained vision-language models trained on billions of visual and language concepts.
Our work aims to produce high-quality geometry outputs by predicting low-frequency shape changes and high-frequency details through source shape \textit{deformations}. We opt to use per-face \textit{Jacobians} as a means for predicting smooth mesh deformations, which enables retaining interesting characteristics of the source shape. This leads to high-quality mesh outputs with useful geometry and mitigates local artifacts commonly caused by vertex displacements. 

We presented a view consistency loss, which avoids over-fitting geometry to specific salient views, and ensures that the same region is roughly interpreted the same from all viewpoints. Our novel loss significantly reduces the number of visual artifacts. We also propose an identity regularization term, which can be controlled by the user to control the magnitude of the deformation. We demonstrate that our method enables retaining interesting global characteristics of the source shape, while still matching it to a highly dissimilar term, providing the user with a controllable and expressive system. 

In the future, we would like to explore the possibility of learning the space of prompt-driven deformations instead of just optimizing them for a single mesh instance. This strategy would be fast, since Jacobians can be predicted in a single feed-forward pass, instead of through optimization. In addition, training over a collection of shapes may induce a neural-regularization which may improve the results further. We would like to also connect our method with a retrieval module to provide a more comprehensive artist-driven creation tool that enables users to explore the results arising from different combinations of sources and prompts.

\bibliographystyle{ACM-Reference-Format}
\bibliography{bibs}

\newpage

\end{document}